\documentclass[letterpaper, 10 pt, conference]{ieeeconf}  

\IEEEoverridecommandlockouts
\usepackage{cite}
\usepackage{amsmath,amssymb,amsfonts}
\usepackage{algorithmicx}
\usepackage{algpseudocode}
\usepackage{algorithm}
\usepackage{graphicx}
\usepackage{subcaption}
\usepackage{textcomp}
\usepackage{xcolor}
\usepackage{afterpage}
\usepackage{tablefootnote}

\usepackage{placeins}
\usepackage{booktabs}
\usepackage{array}
\usepackage{color, colortbl}
\usepackage{romannum}
\usepackage{hyperref}

\usepackage{bm} 
\usepackage[labelfont=bf,textfont=normalfont]{caption}

\definecolor{taupegray}{rgb}{0.55, 0.52, 0.54}
\definecolor{gainsboro}{rgb}{0.86, 0.86, 0.86}

\def\BibTeX{{\rm B\kern-.05em{\sc i\kern-.025em b}\kern-.08em
    T\kern-.1667em\lower.7ex\hbox{E}\kern-.125emX}}

\author{Shijie Fang$^{1^{\dagger}}$, Wenchang Gao$^{1^{\dagger}}$, Shivam Goel$^{1^{\dagger}}$, Christopher Thierauf$^{1}$, Matthias Scheutz$^{1}$, Jivko Sinapov$^{1}$ 
\thanks{\textsuperscript{†}These authors contributed equally to this work, and their names are listed in alphabetical order by last name.}
\thanks{$^{1}$Tufts University School of Engineering, Computer Science. Medford, Massachusetts, United States of America
        \tt\small\{fistname.lastname\}@tufts.edu}
}

\begin{document}

\title{\LARGE \bf FLEX: A Framework for Learning Robot-Agnostic Force-based Skills \\ Involving Sustained Contact Object Manipulation}
\maketitle

\begin{abstract}
Learning to manipulate objects efficiently, particularly those involving sustained contact (e.g., pushing, sliding) and articulated parts (e.g., drawers, doors), presents significant challenges. Traditional methods, such as robot-centric reinforcement learning (RL), imitation learning, and hybrid techniques, require massive training and often struggle to generalize across different objects and robot platforms. 
We propose a novel framework for learning object-centric manipulation policies in \textit{force space}, decoupling the robot from the object. By directly applying forces to selected regions of the object, our method simplifies the action space, reduces unnecessary exploration, and decreases simulation overhead.
This approach, trained in simulation on a small set of representative objects, captures object dynamics—such as joint configurations—allowing policies to generalize effectively to new, unseen objects. Decoupling these policies from robot-specific dynamics enables direct transfer to different robotic platforms (e.g., Kinova, Panda, UR5) without retraining.
Our evaluations demonstrate that the method significantly outperforms baselines, achieving over an order of magnitude improvement in training efficiency compared to other state-of-the-art methods. Additionally, operating in force space enhances policy transferability across diverse robot platforms and object types. We further showcase the applicability of our method in a real-world robotic setting. Link: \url{https://tufts-ai-robotics-group.github.io/FLEX/}
\end{abstract}


\section{Introduction}

Learning to manipulate objects, particularly in tasks involving continuous interaction, such as pushing, sliding, or handling articulated objects like drawers and doors, is a significant challenge in robotics. These tasks, often referred to as sustained contact manipulation~\cite{suomalainen2022survey}, require the continuous application of force throughout the interaction. Efficiently learning such policies is especially difficult when aiming to generalize across different objects and robotic platforms.

Learning-based approaches like reinforcement learning (RL) and imitation learning (IL) typically focus on robot-centric control strategies tailored to specific tasks or configurations. However, these methods require extensive training data and often struggle to generalize to unseen objects or transfer across different robotic systems. Additionally, traditional control techniques such as model predictive control (MPC) rely on precise object and robot dynamics modeling, limiting their applicability in real-world environments.

%
\begin{figure}[t]
    \centering
    \includegraphics[width=0.9\linewidth]{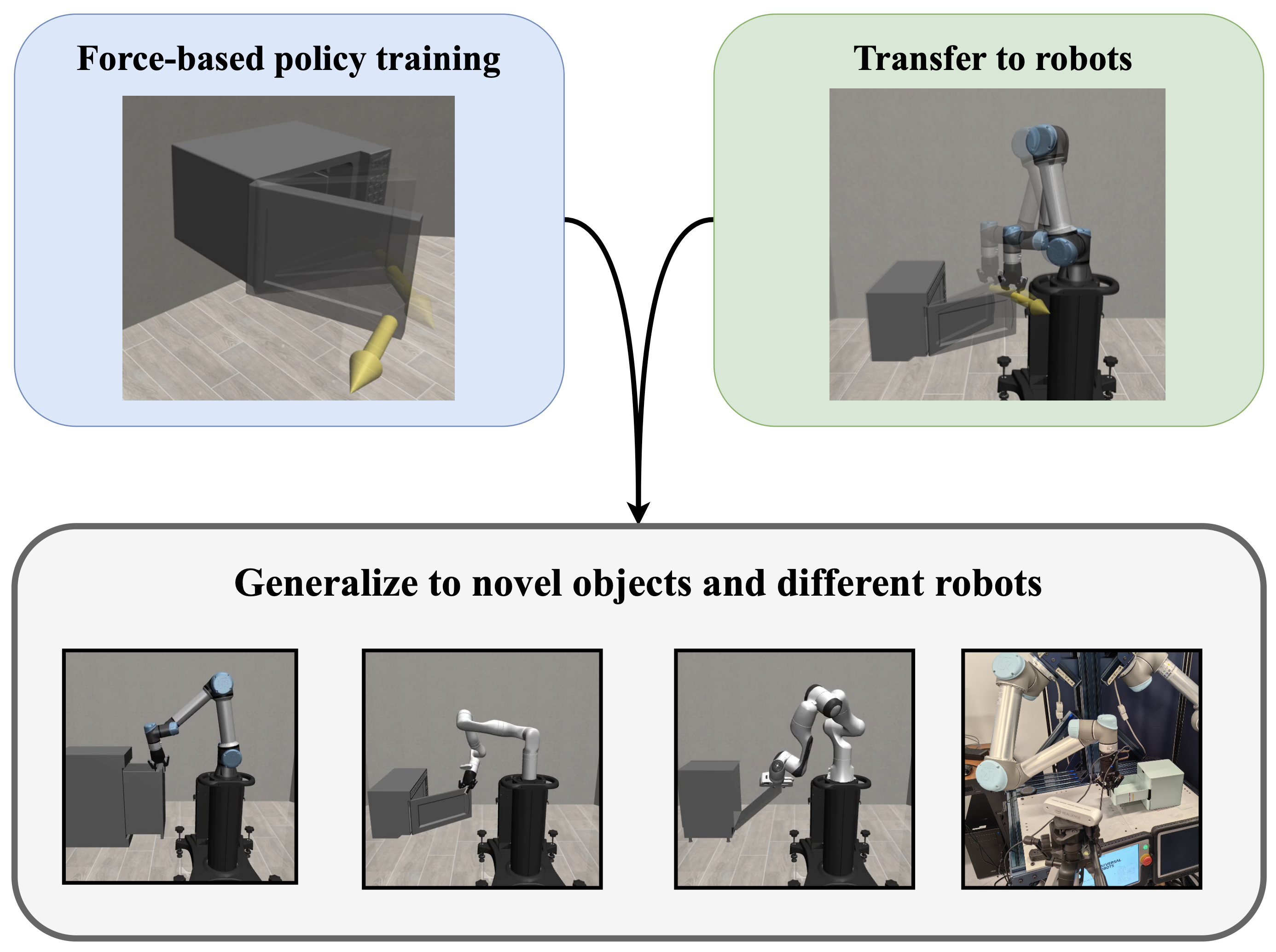}
    \caption{ \small Our method (\textbf{FLEX}: \textbf{F}orce-based \textbf{L}earning for \textbf{EX}tended Manipulation) learns force-based skills for manipulating articulated objects across various robots by decoupling policy learning from robot dynamics. The blue box (top-left) shows the core mechanism of force-based skill learning, while the green box (top-right) illustrates the transfer to different robots. The white box (bottom) highlights manipulation tasks: L-R, the UR5e opens a cabinet, Kinova opens a microwave door, Panda opens a dishwasher rack, and a real UR5 opens a drawer. Detailed architecture in Figure~\ref{fig:overall_diagram}.}
    \label{fig:illustrative_fig}
\end{figure}
%
Recent efforts have sought to simplify RL by leveraging task structures to improve efficiency~\cite{abbatematteo2024composable} and by decoupling object-specific dynamics from robot control to enhance generalization~\cite{lu2023decoupling}. While they show potential, achieving both efficient learning and robust generalization across diverse objects and robots remains challenging, particularly for tasks involving sustained contact and articulated object dynamics.

This work proposes an approach (Figure~\ref{fig:illustrative_fig}) to address these challenges by learning object-centric RL policies in \textit{force space}. Rather than relying on robot-specific control strategies, our method governs object behavior through forces applied directly to the object. By training policies in simulation, we capture key object dynamics, such as movement constrained by joint configurations, which enables generalization across objects of the same type. This allows the learned policies to adapt effectively to novel objects by focusing on their motion-based properties and joint constraints.

Additionally, our method reduces reliance on pre-trained perception models for identifying joint types and configurations. By decoupling policies from low-level robot dynamics, it enables direct transfer across platforms without retraining, while object-centric force interactions improve data efficiency, policy performance, and generalization across diverse objects and robots.

Our method offers three key advantages for sustained contact manipulation of articulated objects: (1) improved data and time efficiency, achieved by representing actions in force space and eliminating unnecessary exploration, which accelerates training by removing robot-specific kinematics from the simulation; (2) generalizability across objects with similar dynamics by capturing essential object features; and (3) direct transferability across different robotic platforms, as demonstrated through evaluations on the Kinova, UR5, and Panda robots in simulation and UR5 arm in real-world.

\section{Related Works}
\noindent \textbf{Robot learning for sustained contact manipulation}:
Sustained contact manipulation, involving tasks like pulling drawers or opening microwaves, poses challenges due to high-dimensional action spaces and extended task duration~\cite{kroemer2021review}. Early RL and IL approaches~\cite{gu2017deep, kalashnikov2018scalable} were hidered by time-consuming real-world data collection. Simulated environments~\cite{todorov2012mujoco, zhu2020robosuite, Xiang_2020_CVPR} have improved efficiency but still struggle with stable interactions throughout extended manipulation tasks and managing complex state-action dynamics~\cite{elguea2023review}.

Recent works simplify RL by reducing action spaces through decoupling high-level policies from low-level motion planning~\cite{xia2021relmogen, nasiriany2022augmenting}, primitive actions~\cite{nasiriany2022augmenting, pertsch2021accelerating}, or robot control~\cite{lu2023decoupling}. Nasiriany et al.\cite{nasiriany2022augmenting} use predefined primitive actions, while Lu et al.\cite{lu2023decoupling} decouple object dynamics from robot control. Other strategies, like structured RL combined with task and motion planning~\cite{rosen2022role} or model-based approaches~\cite{lee2020guided}, aim to reduce the state space but struggle to prevent suboptimal exploration into irrelevant states.

Learning in the force domain offers a promising way to reduce action space dimensionality and prevent unnecessary exploration. Combining force-based RL with symbolic planning has been explored~\cite{thierauf21fixing}, but reliance on accurate digital twins limits generalization to new objects. Similarly, using video demonstrations to apply force on objects~\cite{li2024ag2manip} improves exploration, but guiding the force applier requires significant exploration, increasing training time. Additionally, dependence on large, high-quality demonstration datasets and extensive visual training exacerbates the sim-to-real gap.

\noindent \textbf{Object-centric representation for manipulation} focus on objects rather than specific robots, enabling generalization across tasks and robotic platforms~\cite{kroemer2021review}. Methods leveraging 3D perception, such as object-centric actions~\cite{mo2021where2act}, trajectories~\cite{wu2021vat}, and affordances~\cite{geng2023rlafford}, enhance the transferability of skills to new objects. For example, point clouds can infer joint configurations~\cite{yu2024gammageneralizablearticulationmodeling}, while articulation flow prediction guides action direction~\cite{eisner2022flowbot3d}. UMPNet~\cite{xu2022universal} 
generates closed-loop SE(3) action trajectories from RGB-D inputs using Arrow-of-Time inference, enabling generalization across unseen articulation structures without explicit joint modeling.

While these approaches have simplified robot learning through action space reduction using object-centric representations, generalizing across objects and platforms remains challenging — particularly due to their reliance on pose-space control and robot-specific execution, which limit transferability and hinder stable force application in sustained contact interactions. We build on these efforts by focusing on force-space learning of manipulation skills, targeting object properties like joint configurations. Decoupling robot-specific dynamics and operating solely in object space avoids suboptimal exploration and reduces training time. This accelerates learning, enhances generalization, and enables easy policy transfer across robotic platforms without retraining.

\section{Preliminaries}
\label{sec:prelim}

\subsection{Articulated Object Parametrization}
\label{subsec:object_parameterization}

An articulated object consists of multiple rigid bodies, referred to as \textbf{links}, connected by \textbf{joints} that permit relative motion of object parts. In this work, we focus on objects with joints that allow one degree of freedom, either rotational (revolute) or translational (prismatic). Each joint connects parent and child links, enabling the object to change configuration\cite{eisner2022flowbot3d}. Many real-world articulated objects (e.g., drawers, doors) follow this structure, with their dynamics determined by the joint type and configuration.

\begin{figure}[b]
    \centering
    \begin{subfigure}{0.2\textwidth} 
    \centering
    \includegraphics[width=0.8\textwidth]{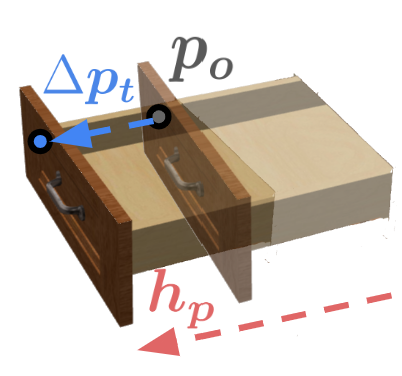}
    \caption{Prismatic}
    \label{fig:prismatic}
    \end{subfigure}
    \hfill
     \begin{subfigure}{0.2\textwidth} 
     \centering
    \includegraphics[width=0.9\textwidth]{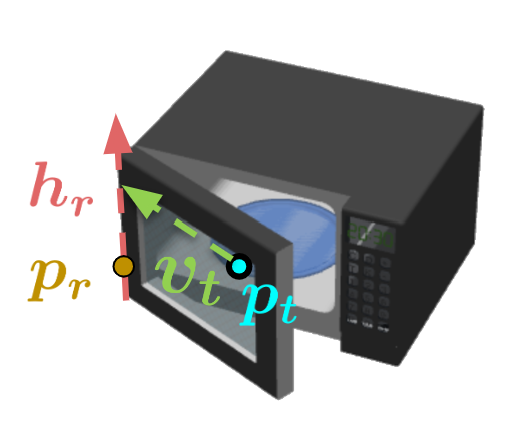}
    \caption{Revolute}
    \label{fig:revolute}
    \end{subfigure}
    \caption{\small Illustration of the two joint configurations and states}
    \label{fig:joint_config_illustration}
\end{figure}

\noindent \textbf{Prismatic joint:} A prismatic joint allows for linear motion along a fixed axis. The position of a point \(\boldsymbol{p}\) along the prismatic joint is defined as:
    \begin{equation}
    \label{eq:prismatic}
     \boldsymbol{p} = \boldsymbol{p_0} + k \cdot \boldsymbol{h_p}, \quad k \in \mathbb{R}, \quad \boldsymbol{p}, \boldsymbol{p_0} \in \mathbb{R}^3
    \end{equation}
    where \(\boldsymbol{p_0}\) is the initial position, \(\boldsymbol{h_p} \in \mathbb{R}^3\) is the unit vector representing the direction of the joint axis, and \(k\) is the scalar displacement along the axis (Figure~\ref{fig:prismatic}).
    
\noindent \textbf{Revolute joint:} A revolute joint allows for rotational movement around a fixed joint axis. The position of a point \(\boldsymbol{p}\) on a revolute joint satisfies the constraint:
    \begin{equation}
    \label{eq:revolute}
        \left\| (\boldsymbol{p} - \boldsymbol{p_r}) - \left[ (\boldsymbol{p} - \boldsymbol{p_r}) \cdot \boldsymbol{h_r} \right] \boldsymbol{h_r} \right\| = r
    \end{equation}
where \(\boldsymbol{p_r} \in \mathbb{R}^3\) is a point on the axis of rotation (joint origin), \(\boldsymbol{h_r} \in \mathbb{R}^3\) is the unit vector along the rotation axis, and $r$ is the radius of the circular motion. This equation constrains \(\boldsymbol{p}\) to move in a circular path around the axis (Figure~\ref{fig:revolute}).

%
\subsection{Reinforcement Learning for Force-Based Manipulation}
\label{sub_sec:rl}
We formulate the manipulation of articulated objects as an RL problem, aiming to learn force-based policies that generalize across objects with similar joint configurations. This problem is modeled using two Markov Decision Processes (MDPs): one for prismatic joints, $M_p = \langle S_p, A, T_p, R, \gamma \rangle$, and one for revolute joints, $M_r = \langle S_r, A, T_r, R, \gamma \rangle$. In this setup, the state spaces $S_p$ and $S_r$ represent the object’s joint-specific information, such as joint axis direction and configuration of the object. The shared action space $A$ consists of 3D force vectors applied to the surface of the object.
The transition functions $T_p$, and $T_r$ govern the object's motion dynamics based on the applied forces, reflecting the behavior of prismatic and revolute joints, respectively. The reward function $R$ encourages efficient manipulation by maximizing object displacement with minimal applied force, with $\gamma$ being the discount factor.
Through this formulation, we aim to learn two policies: $\pi_p$ for prismatic joints and $\pi_r$ for revolute joints, trained to maximize the expected cumulative reward: $G_t = \mathbb{E}\left[\sum_{k=0}^{\infty} \gamma^k R(s_t, a_t, s_{t+1})\right]$, where $s_t \in S_p$ or $S_r$ and $a_t \in A$. These policies will generalize across articulated objects with the same joint configurations, allowing the robot to manipulate a wide range of objects by applying the correct force-based strategy for the given joint type.

\subsection{Problem Formulation}
\label{subsec:problem_formulation}
The challenge we address is solving sustained contact manipulation tasks for articulated objects characterized by prismatic or revolute joints. This requires learning force-based policies in simulation that generalize across various robotic platforms by adapting to specific joint dynamics. Key aspects include (1) learning efficient force-based manipulation strategies for both joint types and (2) executing these learned policies across different robots without retraining.

The problem can be broken down into three components:

\noindent (1) \textbf{Policy learning}: The objective is to learn two manipulation policies: one for prismatic joints, $\pi_p: S_p \rightarrow A$, and one for revolute joints, $\pi_r: S_r \rightarrow A$. These policies must handle different contact points on objects and generalize across various objects with the same joint type. The learned policies should also be directly transferable to different robotic platforms without retraining.

\noindent (2) \textbf{Joint identification during policy execution}: The robot must infer the joint type (prismatic or revolute) of the object it is interacting with. Observing a trajectory of $N$ end-effector positions $\xi = \{\boldsymbol{p_i^{eef}} \in \mathbb{R}^3\}_{i=1}^{N}$, the joint type $J \in \{\text{prismatic, revolute}\}$ is inferred by maximizing the likelihood of the trajectory by Maximum Likelihood Estimation (MLE) problem:
\begin{equation}
\label{eq:MLE}
    \hat{J} = \underset{J \in \{\text{prismatic, revolute}\}}{\operatorname{argmax}}~ p(\xi \mid J)
\end{equation}
where $p(\xi \mid J)$ is the likelihood of trajectory $\xi$ conditioned on joint type $J$ and its parameters (e.g., joint axis, joint point, etc.). The robot selects the joint type with the highest likelihood and applies the corresponding pre-learned policy.

\noindent (3) \textbf{Efficient execution across robots}: The learned object-centric policies generalize across robotic platforms, allowing manipulation without the need for retraining. This ensures that policies learned in an object-centric simulation are directly transferable to different robots.

\section{Learning Methodology}
\label{sec:learning_method}
\begin{figure*}[t]
    \centering
    \includegraphics[width=0.9\textwidth]{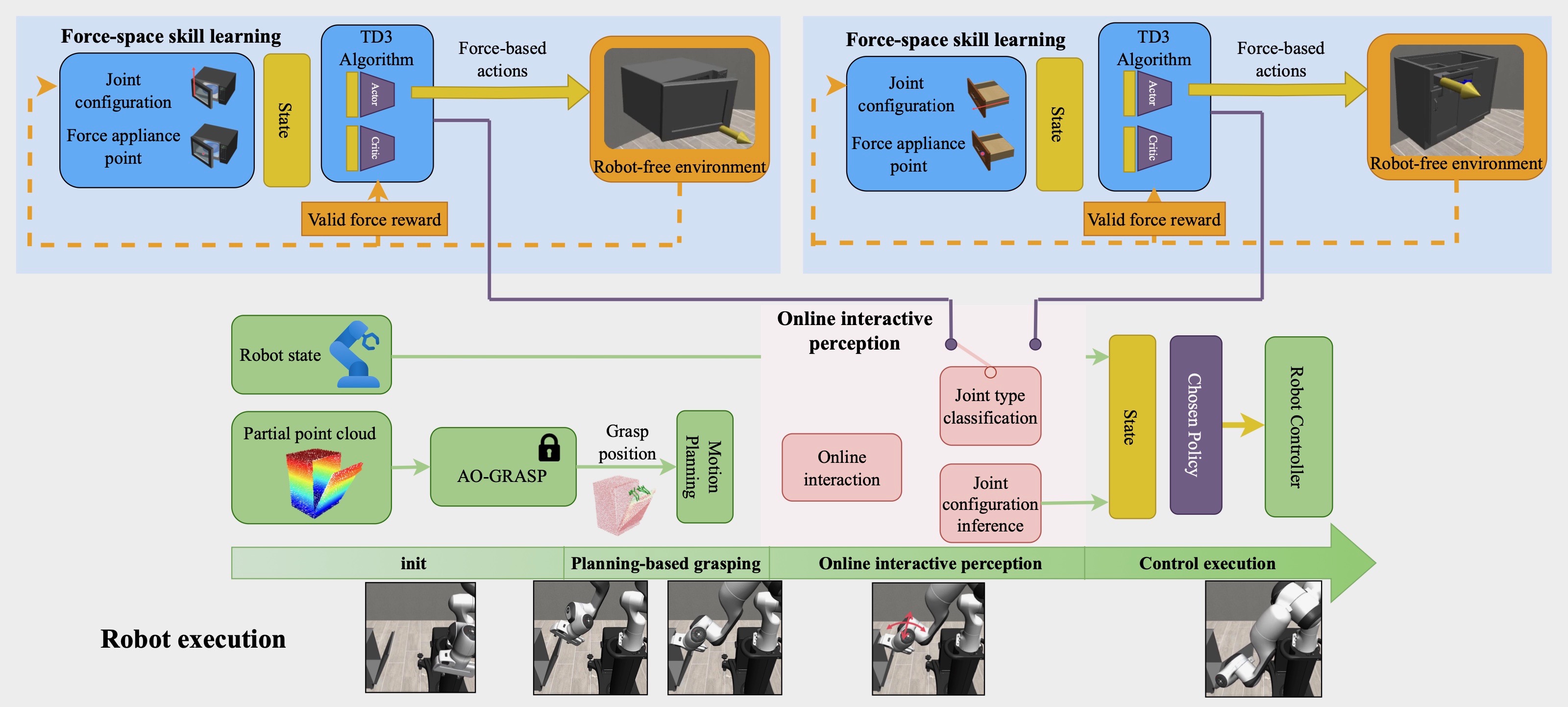}
    \caption{\small The figure presents the architecture of \textbf{FLEX} (\textbf{F}orce-based \textbf{L}earning for \textbf{EX}tended Manipulation) for sustained contact manipulation of articulated objects. The blue boxes represent \textit{Force-based skill learning} in simulation, the red boxes depict \textit{Interactive joint parameter estimation}, and the green boxes show the \textit{Robot execution} phase, where the learned policies are applied to perform tasks.}
    \label{fig:overall_diagram}
\end{figure*}

The architecture of FLEX consists of three main components: (1) Force-space skill learning, (2) Joint parameter estimation, and (3) Robot execution, as shown in Figure~\ref{fig:overall_diagram}.
\subsection{Force-based skill learning}
\label{subsec:skill_learning}
We train object-centric manipulation policies in a simulation environment without a robot. As described in Section~\ref{sub_sec:rl}, learning is modeled using two MDPs: one for prismatic joints, $M_{p} = \langle S_{p}, A, R, T_{p}, \gamma\rangle$, and one for revolute joints, $M_{r} = \langle S_{r}, A, R, T_{r}, \gamma\rangle$. 
The state spaces $S_{p}$ and $S_{r}$ differ depending on the joint type. Formally:

\begin{small}
\[
\begin{array}{cc}
s^{p}_{t} = \begin{bmatrix} \boldsymbol{h_{p}} \\ \Delta \boldsymbol{p_t} \end{bmatrix}, & 
s^{r}_{t} = \begin{bmatrix} \boldsymbol{h_{r}} \\ \boldsymbol{v_t} \end{bmatrix}
\end{array}
\]
\[
\Delta \boldsymbol{p_t} = \boldsymbol{p_{t}} - \boldsymbol{p_0} 
\]
\[
\boldsymbol{v_t} = (\boldsymbol{p_{t}} - \boldsymbol{p_{r}}) - [(\boldsymbol{p_{t}} - \boldsymbol{p_{r}}) \cdot \boldsymbol{h_{r}}] \boldsymbol{h_{r}}
\]
\end{small}

For prismatic objects, the state $s^p_t \in S_p$ consists of the unit vector $\boldsymbol{h_p} \in \mathbb{R}^3$ representing the joint axis direction, and the displacement $\Delta \boldsymbol{p_t}$, which is the difference between the current force application point $\boldsymbol{p_t} \in \mathbb{R}^3$ and initial position $\boldsymbol{p_0} \in \mathbb{R}^3$ at time step $t$ (for illustration refer to Figure~\ref{fig:prismatic}). 

For revolute objects, the state $s^r \in S_r$ includes the unit vector $\boldsymbol{h_r} \in \mathbb{R}^3$ representing the joint axis direction, and $\boldsymbol{v_t}$, the projection of the current force point $\boldsymbol{p_t} \in \mathbb{R}^3$ onto the joint axis. The joint position is represented by $\boldsymbol{p_{r}} \in \mathbb{R}^3$ (\ref{fig:revolute}).

The action space shared across the two MDPs is defined as $A = \{\boldsymbol{F} \in \mathbb{R}^3 \mid \lVert \boldsymbol{F} \rVert \leq \eta \}$, which is the set of force vectors with a maximum magnitude of $\eta$. Every time step $t$ the policy selects an action $a_t \in A$, which serves as the force $\boldsymbol{F}_t$ applied at the point of contact, and $\eta$ is the maximum allowable force that can be applied to the object by the robot.

The reward function is shared between the two MDPs. It is defined as a dense reward \( R(s_t, a_t, s_{t+1}) \) that encourages efficient manipulation by maximizing object displacement with minimal applied force, as well as a large positive reward when the agent successfully reaches the goal state. Formally,


\begin{scriptsize}
\[
\begin{aligned}
R(s_t, a_t, s_{t+1}) = 
\begin{cases} 
\lVert \Delta x \rVert \cos(\theta_{a_t,\Delta x}) + \mathbb{I}(s_{t+1} = s_{\text{goal}}) R_{\text{goal}} & \text{if } \Delta q \geq 0, \\
-\lVert \Delta x \rVert \cos(\theta_{a_t,\Delta x}) & \text{else}.
\end{cases}
\end{aligned}
\]
\end{scriptsize}

\noindent where: \( \theta_{a_t,\Delta x} \) is the angle between the applied force \( a_t \) and the object’s movement direction \( \Delta x \), \( \Delta q \) represents the change in the object's joint configuration (either position for prismatic joints or rotation angle for revolute joints), \( \mathbb{I}(s_{t+1} = s_{\text{goal}}) \) is an indicator function that equals 1 if the agent reaches the goal state \( s_{\text{goal}} \), and 0 otherwise, and \( R_{\text{goal}} \) is the large positive reward given when the agent successfully reaches the goal state. The transition functions $T_{p}$ and $T_{r}$ model the object’s motion dynamics for prismatic and revolute joints, respectively.

The policies $\pi_r$ (revolute) and $\pi_p$ (prismatic) are learned using TD3~\cite{fujimoto2018addressing}, with each policy network predicting force direction, $\pi_{\text{direction}}(s_t)$, and magnitude, $\pi_{\text{scale}}(s_t) \in [0, 1]$. The applied force at time $t$ is $a_t = \pi_{\text{direction}}(s_t) \cdot \pi_{\text{scale}}(s_t) \cdot \eta$. This separation improves control precision and learning efficiency in force-space manipulation.


The learned policies are contact-agnostic, with contact points randomly sampled from a predefined region in each episode (see Section~\ref{subsec:experimental_setup} for more details). By applying forces at various points, the robot learns a general manipulation strategy, ensuring robustness and adaptability regardless of the exact contact point.

At the end of training, the robot has two learned policies: $\pi_r$ for revolute joints and $\pi_p$ for prismatic joints. The next step is identifying the objects's joint type so the robot can select and apply the correct manipulation. We detail this process of joint estimation in the following subsection.

\subsection{Interactive Joint Parameter Estimation}
\label{subsec:joint_estimation}
We now describe how the robot leverages these skills for object manipulation. The robot platform has an RGB-D camera that generates partial point clouds of the object. The pre-trained AO-GRASP model~\cite{morlans2023grasp} identifies suitable grasp points, and motion planners guide the robot to reach and maintain a successful grasp.

The robot is not provided with any prior information about the object’s joint configuration. To infer the joint type \( J \in \{\text{prismatic, revolute}\} \), it securely holds the object and applies small forces in a specific sequence of predetermined directions—forward, backward, left, and right. This controlled interaction generates a short trajectory $\xi = \{\boldsymbol{p_i^{eef}} \in \mathbb{R}^3\}_{i=1}^{N}$, where each $\boldsymbol{p_i^{eef}}$ is the end-effector position at time step $i$, while ensuring that the robot maintains stable contact with the object. Based on this trajectory, the joint type and its parameters are estimated, guiding the selection of the appropriate manipulation policy.

As introduced in Section~\ref{subsec:problem_formulation}, the problem of joint type inference is formalized as an MLE problem (see equation~\ref{eq:MLE}).

For a prismatic joint, the joint axis $\boldsymbol{\hat{h}_p}$ and origin $\boldsymbol{\hat{p}_o}$ are:
\[
\boldsymbol{\hat{h}_p}, \boldsymbol{\hat{p}_o} = \underset{\boldsymbol{h_p}, \boldsymbol{p_o}}{\operatorname{argmax}}~ p \left(\xi \mid \boldsymbol{h_p}, \boldsymbol{p_{o}}, \text{prismatic}\right)
\]
where $\boldsymbol{h_p}$ is the prismatic joint axis direction, and $\boldsymbol{p_o}$ is the origin of the prismatic joint. The likelihood \( p(\xi \mid \text{prismatic}) \) models the trajectory as linear motion along the joint axis.

Similarly, for revolute joints, the joint axis $\boldsymbol{\hat{h}_r}$, joint point $\boldsymbol{\hat{p}_r}$, and radius $\hat{r}$ are estimated as:
\[
\boldsymbol{\hat{h}_r}, \boldsymbol{\hat{p}_r}, \hat{r} = \underset{\boldsymbol{h_r}, \boldsymbol{p_r}, r \leq \rho}{\operatorname{argmax}}~ p \left(\xi \mid \boldsymbol{h_r}, \boldsymbol{p_{r}}, r, \text{revolute}\right)
\]
where $\boldsymbol{h_r}$ is the joint axis direction, $\boldsymbol{p_r}$ is the joint point, and $r$ is the radius of the circular motion. The constraint \( r \leq \rho \) ensures physical plausibility, where \( \rho \) is the maximum allowable radius.

To estimate these parameters, Principal Component Analysis (PCA) is applied to the end-effector trajectory $\xi$ to yield initial estimates of the joint direction and orientation. These estimates are refined by solving least squares optimization.

For prismatic joints, the objective is to minimize the projection errors of the trajectory points onto the joint axis:

\[
\underset{\boldsymbol{p_o}, \boldsymbol{h_p}}{\min} \sum_i \left\|\boldsymbol{p_i^{eef}} - \boldsymbol{p_o} - \left[\left(\boldsymbol{p_i^{eef}} - \boldsymbol{p_o}\right) \cdot \boldsymbol{h_p}\right] \boldsymbol{h_p}\right\|_2^2
\]
This ensures that the trajectory points lie along the straight path, characteristic of prismatic motion.

For revolute joints, the goal is to minimize the error between the trajectory points and a circle of radius $r$ around the joint point, as defined by:

\[
\underset{\boldsymbol{p_r}, r \leq \rho}{\min} \sum_i \left(\left\|\boldsymbol{p_i^{eef}} - \boldsymbol{p_r}\right\|_2^2 - r^2\right)
\]
This captures the revolute motion, where the object’s movement follows a circular trajectory.
After determining the joint parameters, the robot uses the observed trajectory to compute the likelihood for both joint types. The robot then compares the likelihoods of the prismatic and revolute models, selecting the joint type \( J \) that maximizes the likelihood:
\[
\hat{J} = \underset{J \in \{\text{prismatic, revolute}\}}{\operatorname{argmax}}~ p(\xi \mid J, \hat{\boldsymbol{\theta}_J}),
\]
where \( \hat{\boldsymbol{\theta}_J} \) represents the estimated parameters for joint type $J$, such as the axis \(\boldsymbol{h}\) and origin \(\boldsymbol{p_o}\) for prismatic joints, or the axis \(\boldsymbol{h_r}\), point \(\boldsymbol{p_r}\), and radius $r$ for revolute joints.

\subsection{Robot Execution}
\label{subsec:robot_execution}
The final step applies the learned force-based manipulation policies to robotic platforms. The execution pipeline (green in Figure~\ref{fig:overall_diagram}) consists of three components: perception and grasping, state construction, and control execution.

\noindent \textbf{Perception and grasping}:  
The robot uses its onboard RGB-D camera to generate partial point clouds of the object. Using the pre-trained AO-GRASP model~\cite{morlans2023grasp}, it identifies optimal grasp points, and a motion planner helps securely grasp the object, ensuring stable interaction before manipulation.

\noindent \textbf{State construction}:  
After grasping the object, the robot uses the interactive joint parameter estimation (Section~\ref{subsec:joint_estimation}) to infer the joint type and configuration \( [\boldsymbol{\hat{h}_J}, \boldsymbol{\hat{p}_J}] \). This knowledge, along with the robot’s end-effector position \( \boldsymbol{p_i^{eef}} \), is used to construct the state for the manipulation policy. This object-centric state construction is crucial, as it enables \textit{hard transfer} by aligning the policy learned in simulation with the object’s real-world configuration, allowing direct application of the learned policies without further retraining.

\noindent \textbf{Control execution}:  
Once the state is constructed, the robot selects the policy \( \pi_J \), which provides the forces to be applied based on the object’s joint dynamics. The forces predicted by the policy are scaled and mapped to generate the desired end-effector (EEF) pose changes. These pose changes are then converted into joint torques via an Operational Space Controller (OSC)~\cite{khatib1987unified}, ensuring smooth adaptation to variations in object weight and dynamics. The robot applies the specified forces until the manipulation task (e.g., opening a drawer or microwave door) is successfully completed.

\noindent The learned policy does not dynamically adjust force magnitude for unforeseen factors like friction or spring tension. Addressing this limitation is left for future work, particularly for handling novel objects with varying dynamics.

\section{Experiments}
We evaluate FLEX on three key aspects: learning efficiency in simulation, policy transferability across objects, and its task execution performance across different robots in simulation and in the real world.

\subsection{Experimental Setup}
\label{subsec:experimental_setup}

The experiments were conducted in Robosuite~\cite{zhu2020robosuite}, a simulation platform built on Mujoco, focusing on four manipulation tasks: opening cabinets, trashcans, microwaves, and dishwashers. Cabinets and dishwashers have prismatic joints, while microwaves and trashcans involve revolute joints. All objects are sourced from the PartNet-Mobility dataset~\cite{xiang2020sapien}.


We trained two force-based policies: one for prismatic joints (using a drawer) and another for revolute joints (using both a microwave and a dishwasher\footnote{Two objects were used due to the complexity of this joint type.}). For revolute joints, either the microwave or dishwasher is randomly chosen at the start of each episode. During training, the agent is provided with ground-truth joint configurations and randomized force application points from accessible regions, such as handles and front panels, derived from the objects’ URDF files.

Training efficiency is measured through wall time and timesteps. The simulation runs at $20$ Hz, with each timestep corresponding to $0.05$ seconds of simulated time. A convergence criterion is defined based on the alignment between the applied force and the object’s movement direction:
$\chi = \frac{1}{N}\sum_{t=0}^N \frac{a_{t} \cdot \Delta x }{\lVert a_{t} \rVert \lVert \Delta x \rVert} = \frac{1}{N}\sum_{t=0}^N\cos{\theta_{a_t,\Delta x}}$, where $\chi$ is the average percentage of force applied in the correct direction throughout the episode. Once $\chi$ averages at least $0.75$ over the last $100$ episodes, the policy is considered to have converged. This criterion yields a $100\%$ success rate for both prismatic and revolute joint scenarios during training\footnote{All experiments, including baselines, were run on Intel® Core™ i9-13900 (5 GHz) and RTX 4090 GPU, with 24 processes for multi-process.}.

For force-based reinforcement learning, each training episode has 200 timesteps, while robot-centric reinforcement learning baselines extend to 500 steps per episode. Our method reduces the required timesteps and accelerates simulation by excluding the robot from the environment, thereby eliminating computational costs associated with collisions, physics, and robot control. 
The training process is conducted across ten independent trials, with the mean and standard deviation of the training times in Table~\ref{tab:results} (columns 1\& 2).



\subsection{Evaluation and Metrics}

We evaluate FLEX's ability to generalize to unseen objects and complete the manipulation task autonomously. The evaluation is conducted using 13 previously unseen objects from the PartNet-Mobility~\cite{xie2023part} dataset, including four microwaves, three dishwashers, three trashcans, and three cabinets. For each object, the robot is provided with a partial point cloud and must autonomously (1) infer the joint type, (2) grasp the object using AO-GRASP~\cite{morlans2023grasp} and motion planner, (3) load the corresponding pre-trained policy, and (4) execute the manipulation skill. Success is defined as opening the object to at least $80\%$ of its joint limit while maintaining a stable grasp\footnote{Relaxed to 70\% for baselines to avoid zero success rates.}. To prevent “lucky success,” any incorrect joint classification leads to immediate task failure.

To streamline the evaluation process, grasp proposals from the AO-GRASP model are generated once at the beginning of each trial and remain consistent throughout the 20 manipulation attempts for each object. The evaluation protocol is repeated for all 13 unseen objects, with 20 trials conducted per object. The success rate is computed for each class of objects, and we report the mean and standard deviation for 10 independent trials, as shown in Table~\ref{tab:results} (column 3).

We also assess the transferability of the learned policies across different robots (Panda, UR5e, and Kinova Gen3) within the Robosuite environment, recording success rates for each manipulation task. To demonstrate the real-world applicability of our method, we conduct a proof-of-concept experiment using a UR5 robotic arm to open a drawer.\footnote{Video demonstration is available in the supplementary materials.}

\subsection{Baselines}
\label{sucsec:baselines}

\noindent\textbf{End-to-end RL:}
We implemented an end-to-end RL baseline using the TD3 algorithm from Stable-Baselines3~\cite{raffin2021stable}
with a hand-crafted, dense, staged reward function inspired by \cite{lu2023decoupling}.

\noindent\textbf{CIPS:}
We selected CIPS~\cite{rosen2022role}, a hybrid RL and planning method that reduces task horizon by planning up to the point of object contact. 
To speed up training, the planning phase was skipped by pre-recording 4 instances per object, with the robot already grasping the object at the start of each episode. 


\noindent\textbf{GAMMA:}
To benchmark against a SOTA visual perception-based manipulation method, we selected GAMMA~\cite{yu2024gammageneralizablearticulationmodeling}, a pre-trained model that infers joint configurations and grasp poses from partial point clouds. As GAMMA is a pre-trained model that uses planning, training time was not applicable.

All RL-based baselines were trained on the same objects and ground-truth observations as FLEX. During rollout, we provide ground-truth joint configuration and type for these baselines. We assume the ground-truth knowledge for joint type is needed to complete the task, and we select the correct result from multiple proposals given by GAMMA. To maintain consistency, we use AO-GRASP for grasp proposals in both CIPS and GAMMA.

\begin{table}[t]
\footnotesize
\begin{center}

\begin{tabular}{>{\centering}p{1.8cm}>{\centering}p{1.85cm}>{\centering}p{1.85cm}p{1.3cm}}
\arrayrulecolor{taupegray}\toprule

\textbf{\begin{small}Algorithm\end{small}} & {\textbf{\begin{small}Training (timesteps)\end{small} }}  & {\textbf{\begin{small}Training (walltime)\end{small} }}& \textbf{\begin{small}Success Rate\end{small} }
\tabularnewline
\midrule
 &\begin{footnotesize} Mean$\pm$SD\end{footnotesize}
 & \begin{footnotesize}Mean$\pm$SD\end{footnotesize}
  & \begin{footnotesize}Mean$\pm$SD\end{footnotesize}
\tabularnewline

\midrule
 \multicolumn{4}{c}{\textbf{Cabinet [Prismatic]}} \tabularnewline
 \midrule

\rowcolor{gainsboro}
FLEX  & $0.4 \pm 0.26$ M & $0.36 \pm 0.2$ h & \centering $0.91 \pm 0.08$ 
\tabularnewline 
End-to-end RL  & $\geq 10$ M & $\geq 24$ h & \centering $\leq 0.01$ 
\tabularnewline
CIPS  & 3 M & $1.46 \pm 0.02$ h & \centering $0.34 \pm 0.31$
\tabularnewline
GAMMA+Planning & N/A & N/A & \centering $0.49 \pm 0.29$
\tabularnewline

\midrule
 \multicolumn{4}{c}{\textbf{Trashcan [Prismatic]}} \tabularnewline
 \midrule

\rowcolor{gainsboro}
FLEX  & $0.4 \pm 0.26$ M & $0.36 \pm 0.2$ h & \centering $0.88 \pm 0.10$ 
\tabularnewline
End-to-end RL  & $\geq 10$ M & $\geq 24$ h & \centering $\leq 0.01$ 
\tabularnewline
CIPS  & 3 M & $1.46 \pm 0.02$ h & \centering $0.19 \pm 0.25$ 
\tabularnewline
GAMMA+Planning & N/A & N/A & $0.22 \pm 0.18$
\tabularnewline

\midrule
 \multicolumn{4}{c}{\textbf{Microwave [Revolute]}} \tabularnewline
 \midrule

\rowcolor{gainsboro}
FLEX  &  $0.9 \pm 0.3$  M & $0.63 \pm 0.22$ h &\centering $0.67 \pm 0.14$
\tabularnewline

End-to-end RL  & $\geq10$ M & $\geq 24$ h &\centering $\leq 0.01$
\tabularnewline
CIPS  & 5 M & \centering $2.84 \pm 0.2$ h & \centering $0.33 \pm 0.26$
\tabularnewline
GAMMA+Planning & N/A & N/A & \centering $0.49 \pm 0.18$
\tabularnewline

\midrule
 \multicolumn{4}{c}{\textbf{Dishwasher [Revolute]}} \tabularnewline
 \midrule

\rowcolor{gainsboro}
FLEX  & $0.9 \pm 0.3$ M & $0.63 \pm 0.22$ h & \centering $0.64 \pm 0.11$ 
\tabularnewline

End-to-end RL  & $\geq 10$ M & $\geq 24$ h & \centering $\leq 0.01$ 
\tabularnewline
CIPS  & 5 M & $2.84 \pm 0.2$ h & \centering $0.55 \pm 0.36$ 
\tabularnewline
GAMMA+Planning & N/A & N/A & \centering $0.47 \pm 0.38$
\tabularnewline

\bottomrule
\end{tabular}
\caption{\small Results on novel object instances evaluated using the Panda robot. Our method (FLEX) is highlighted in grey. Training time in hours, and timesteps in millions.}
\label{tab:results}
\end{center}
\end{table}

\section{Results \& Discussion}
\label{sec:results_discussion}

The results of our experiments are summarized in Tables~\ref{tab:results} and \ref{tab:results_portrait}. It can be seen that FLEX consistently outperforms all baselines, including CIPS~\cite{abbatematteo2024composable} and GAMMA~\cite{yu2024gammageneralizablearticulationmodeling}, with success rates ranging from $64\%$ to $91\%$ across various objects. In some cases, FLEX achieves up to four times higher success rates, demonstrating its superior performance in task execution. Additionally, FLEX shows significantly lower standard deviations in success rates, emphasizing the stability and reliability of the learned policies.

FLEX significantly reduces training time, converging faster with fewer timesteps and less wall time, as shown in Table~\ref{tab:results}. This efficiency is twofold: by decoupling the robot from the training environment, FLEX avoids simulating robot dynamics, collisions, and physics, accelerating training and reducing computational costs. Additionally, learning in force space reduces the action space and prevents inefficient exploration (e.g., detaching from objects), allowing FLEX to converge to a successful policy with fewer timesteps and improved sample efficiency compared to the baselines.





\begin{table}[t]
\footnotesize
\begin{center}
\begin{tabular}{>{\centering\arraybackslash}p{2.5cm}>{\centering\arraybackslash}p{1.3cm}>{\centering\arraybackslash}p{1.5cm}>{\centering\arraybackslash}p{1.5cm}}
\toprule

\textbf{Object} & \textbf{Panda} & \textbf{UR5e} & \textbf{Kinova3} \tabularnewline
\arrayrulecolor{taupegray}\midrule

\textbf{Cabinet} &  $0.91 \pm 0.08$ & $0.82 \pm 0.13$ & $0.77 \pm 0.12$ \tabularnewline
\arrayrulecolor{gainsboro}\midrule
\textbf{Trashcan} &  $0.88 \pm 0.10$ & $0.75 \pm 0.15$ & $0.79 \pm 0.10$ \tabularnewline
\midrule
\textbf{Microwave} &  $0.67 \pm 0.14$ & $0.53 \pm 0.08$ & $0.55 \pm 0.12$ \tabularnewline
\midrule
\textbf{Dishwasher} &  $0.64 \pm 0.11$ & $0.62 \pm 0.11$ & $0.59 \pm 0.13$ \tabularnewline

\arrayrulecolor{taupegray} \bottomrule
\end{tabular}
\caption{\small Success rates for different objects across different robotic platforms.}
\label{tab:results_portrait}
\end{center}
\end{table}

In contrast, the end-to-end RL baseline struggles to perform the task, even after 10 million timesteps of training, achieving negligible success rates despite using a staged reward function~\cite{lu2023decoupling}. This highlights the difficulty of the task, particularly in balancing exploration and learning in high-dimensional action spaces -- in robot-centric environments.


CIPS~\cite{rosen2022role} performs moderately well, but inefficient exploration limits its stability. Once the robot enters undesirable states (e.g., detaching from the object), it tends to repeat these behaviors, resulting in poor performance in some trials.

GAMMA performs decently on most object types but struggles with the \textit{trashcan} task, where it fails to infer the correct joint type. This leads to errors in joint configuration inference and subsequent control, highlighting its difficulty in generalizing to unseen objects outside its training set.


A key advantage of FLEX is its transferability across robots. By decoupling manipulation skills from robot-specific dynamics, FLEX can transfer to various robots (Panda, UR5e, Kinova Gen3) without retraining. As shown in Table~\ref{tab:results_portrait}, FLEX performs consistently across robots with minimal degradation, whereas CIPS and GAMMA require significant tuning and adaptation, limiting their scalability.

\section{Conclusion \& Future Work}
We introduced FLEX, a framework for force-based learning of manipulation skills that decouples robot-specific dynamics from policy learning. FLEX enhances training efficiency, reduces action space complexity, and generalizes well across various articulated objects and robotic platforms. Our evaluations demonstrate FLEX’s high success rates and consistent performance across multiple robot platforms, including transfer to real-world settings, as evidenced by a UR5 robot manipulating a drawer.

Future work will address the current lack of dynamic force adjustments in response to varying object properties, such as weight, friction, or joint damping. 
Another promising direction will involve FLEX autonomously generating task-specific environments in real-time, allowing it to adapt and learn policies for novel goals. This would further enhance its effectiveness in open-world scenarios. Finally, integrating learning from demonstrations could improve data efficiency, making the framework more adaptable and robust.

\section*{Acknowledgments}
We thank the reviewers for their valuable feedback and suggestions. We also thank Brennan Miller-Klugman for his valuable help in setting up the robots. This work was supported in part by ONR grant N00014-24-1-2024.

\bibliographystyle{IEEEtran}
\bibliography{root}
\newpage
\appendix

\title{Appendix}

This appendix is organized into several sections to provide comprehensive details about our experimental setup, methodology, and additional results. We cover key aspects such as simulation environments, learning processes, execution details, baseline settings, real-world experiments, and object details. Each section delves into assumptions, settings, hyperparameters, and supplementary results that could not be included in the main paper.

\subsection{Simulation Details}
Our experiments are performed in three simulation environments:

\begin{itemize}
    \item \textbf{TrainingEnv:} This environment excludes the robot and focuses on force-based interactions with objects. Random objects from the training set are selected and placed at \([-1, 0]\) with randomized rotations around the Z-axis. Force application points are randomly sampled, and actions are applied using the simulator. 
    \item \textbf{BaselineTrainingEnv:} Includes both a Panda robot and objects. The objects are placed on a partial circle around the robot to ensure reachability. Object orientations are randomized for evaluation consistency.
    \item \textbf{EvalEnv:} Both FLEX and baseline methods are evaluated in this environment with consistent settings.
\end{itemize}

\subsection{Learning Process}

\subsubsection{Network Design}
The actor network for TD3 has two heads:

\[
\begin{aligned}
\text{Action Direction: } & \text{3 dimensions}, \\
\text{Action Scale: } & \text{1 dimension}.
\end{aligned}
\]
Both the action dimensions and action scale are limited to (0, 1] using bounded activation functions.
The action is then computed as:

\[
a_t = \pi_{\text{direction}}(s_t) \cdot \pi_{\text{scale}}(s_t) \cdot \eta
\]

where \(\eta\) is the maximum allowable force.

\subsubsection{Curriculum Learning for Revolute Objects}
Training for revolute objects uses the following curricula with increasing rotation randomization:
\[
\begin{gathered}
(-0.25, 0.25), \\
(-\pi/2, 0), \\
(-\pi/2, \pi/2), \\
(-\pi, 0), \\
(-\pi, \pi).
\end{gathered}
\]
The next curriculum is used when the agent applies 80\% of force in the correct direction over 100 episodes. 
\subsubsection{Network Structure}
The actor network is a two-headed MLP network. One input layer and two hidden layers encode the state. The scale head and direction head then map the encoded features to normalized three-dimensional action direction and one-dimensional action scale in order to compute actions.

\begin{table}[htbp]
\centering
\begin{tabular}{cc}
\toprule
\textbf{Parameter} & \textbf{Value} \\
\midrule
Actor Hidden layers & 2 \\
Actor Hidden layer dimensions & (400,300) \\
 $\pi_{\text{direction}}$ Activation function & Sigmoid \\
  $\pi_{\text{scale}}$ Activation function & Tahn \\
Critic Hidden Layers & 2 \\
Critic Hidden layer dimensions & (400,300) \\
Critic Activation Function & ReLU \\

\bottomrule
\end{tabular}
\caption{Agent Network Struture}
\end{table}

\subsubsection{RL Hyperparameters}

\begin{table}[t]
\centering
\begin{tabular}{cc}
\toprule
\textbf{Parameter} & \textbf{Value} \\
\midrule
Discount Factor (\(\gamma\)) & 0.99 \\
Batch Size & 100 \\
Learning Rate & 0.001 \\
Policy Update (\(\tau\)) & 0.995 \\
Noise Clip & 1 \\
Policy Delay & 2 \\
Max Timesteps per Episode & 200 \\
Rollouts & 5 \\
State Dimension & 6 \\
Action Dimension & 3 \\
Max Action & 5 \\
 max force magnitude ($\eta$) & $0.02$\\
\bottomrule
\end{tabular}
\caption{Reinforcement Learning Hyperparameters}
\end{table}

The training was performed with 24 parallel environments on an Intel i9-12900K processor with an Nvidia RTX 3080 GPU.

\subsection{Execution Details}

\subsubsection{Execution Assumptions and Object Placement}
To avoid collisions, revolute objects are placed $0.7$m away with a randomized rotation angle between \([- \pi/2, 0]\), while prismatic objects are placed $0.9$m away with an angle between \([-3\pi/4, \pi/4]\).

\subsection{Baselines}

\subsubsection{End-to-End RL}
The end-to-end RL agent is trained using the StableBaselines3 TD3 agent, with objects placed at a \textit{fixed} location. The reward function is designed to guide the robot through sequential stages: approaching, grasping, and manipulating the object. . Despite the use of a dense reward function, the poor (almost nil) performance highlights the complexity of the task and demonstrates why \textit{conventional} RL-based control is an inefficient approach.

\subsubsection{CIPS Training}
To improve efficiency, CIPS training begins with the robot \textit{already} grasping the object. Four pre-recorded initial states are uniformly sampled from the $1/4$ circle around the robot. This method accelerates training by reducing unnecessary exploration.
\subsubsection{GAMMA}
As stated in the paper, we assume knowledge of joint type for GAMMA evaluation. We designed a simple planning method that complies with the ideal joint dynamics. Specifically, for revolute objects, 

\[
\boldsymbol{a}_t^{GAMMA}=-\hat{\boldsymbol{h}_r} \times \hat{\boldsymbol{v}_t}
\]

For prismatic objects, 

\[
\boldsymbol{a}_t^{GAMMA}=
\hat{\boldsymbol{h}_p}
\]

Where $\hat{\boldsymbol{h}_r}$ and $\hat{\boldsymbol{h}_p}$ are the selected prediction results of joint parameters given by GAMMA.

We observed that GAMMA tends to overestimate the number of joints on complex objects with multiple moving parts, leading to incorrect joint inferences. To mitigate this, we only select the first inference result that complies with our ground-truth knowledge of the joint type. If no correct inference exists, we randomly choose from the remaining detected joints.

\begin{table}[t]
\centering
\begin{tabular}{cc}
\toprule
\textbf{Parameter} & \textbf{Value} \\
\midrule
Policy Learning Rate & Default (SB3) \\
Discount Factor (\(\gamma\)) & Default (SB3) \\
Replay Buffer Size & Default (SB3) \\
\bottomrule
\end{tabular}
\caption{Baseline Hyperparameters for CIPS}
\end{table}

\subsection{Real-world setup}

\subsubsection{Hardware and simplifications}
The real-world experiments used a UR-5 arm and an Intel Realsense D455 RGBD camera. To simplify, we used an inverse kinematics (IK) controller and a toy drawer. We assumed knowledge of the drawer pose and skipped interactive perception. The hyperparameters stayed the same with simulation experiment settings.
\subsection{Additional experiments and results}

\subsubsection{80\% opening criterion}
In our experiments, success is defined as opening the object to at least 80\% of its joint limit.

\subsubsection{Additional experiments}
Several additional experiments were conducted on objects with different joint configurations, but due to space limitations, they are not included in the main paper.

\subsection{Object details and dataset}

\subsubsection{Object scaling and preprocessing}
Objects were selected from the Partnet-Mobility dataset and scaled for simulation. Prismatic objects were scaled to 50\% and revolute objects to 30\% of their original size.

\subsubsection{Collision handling}
Due to self-collision issues in the dataset, we disabled collisions for the main body of objects during training. Since the objects were partially opened before manipulation, this did not affect the results.

\end{document}